\documentclass{article} 
\usepackage{iclr2025_conference,times}

\usepackage{amsmath,amsfonts,bm}

\def\eqref#1{equation~\ref{#1}}

\def\1{\bm{1}}

\DeclareMathAlphabet{\mathsfit}{\encodingdefault}{\sfdefault}{m}{sl}
\SetMathAlphabet{\mathsfit}{bold}{\encodingdefault}{\sfdefault}{bx}{n}

\usepackage{hyperref}
\usepackage{url}

\usepackage{algorithm}
\usepackage{algorithmic}
\usepackage{amsmath}
\usepackage{graphicx}

\usepackage{booktabs}
\usepackage{multirow}
\usepackage{soul,color}
\usepackage{xspace}
\usepackage{xcolor}
\usepackage{colortbl}
\usepackage{longtable}
\usepackage[most]{tcolorbox}

\title{A Mamba Foundation Model for Time Series Forecasting}

\author{%
	Haoyu Ma, Yushu Chen \thanks{First Author and Second Author contribute equally to this work. Email: chenyushu@mail.tsinghua.edu.cn}, Wenlai Zhao, Jinzhe Yang \thanks{Also at National Supercomputing Center and Tecorigin in Wuxi, Jiangsu, China.}, \\
	Department of Computer Science and Technology\\
	Tsinghua University\\
	Beijing, China, 100084\\
        \AND
        Yingsheng Ji \\ 
        Peng Cheng Laboratary \\
        Shenzhen, China, 518000\\
	\AND
	Xinghua Xu \\
	Naval University of Engineering,\\
	Wuhan, China, 430033\\	
	\AND
	Xiaozhu Liu \\
	Beijing Institute of Technology\\
	Beijing, China, 100081\\
 	\AND
	Hao Jing \\
	Earth System Modeling and Prediction Center \\
	China Meteorological Administration \\
	Beijing, China, 100081\\
	\AND
	Shengzhuo Liu \\
	College of Computer Science and Mathematics\\
	Fujian University of Technology\\
	Fuzhou, China, 350118\\
	\AND
	Guangwen Yang \thanks{Corresponding author. Also at National Supercomputing Center in Wuxi, Jiangsu, China, and Zhejiang Lab, Hongzhou, China. }\\
	Department of Computer Science and Technology\\
	Tsinghua University\\
	Beijing, China, 100084\\
}

\newcommand{\boldres}[1]{{\textbf{\textcolor{red}{#1}}}}
\newcommand{\secondres}[1]{{\underline{\textcolor{blue}{#1}}}}

\definecolor{tabhighlight}{HTML}{e5e5e5}

\iclrfinalcopy 
\begin{document}

	\maketitle
	
	\begin{abstract}
		
		Time series foundation models have demonstrated strong performance in zero-shot learning, making them well-suited for predicting rapidly evolving patterns in real-world applications where relevant training data are scarce. However, most of these models rely on the Transformer architecture, which incurs quadratic complexity as input length increases. To address this, we introduce TSMamba, a linear-complexity foundation model for time series forecasting built on the Mamba architecture. The model captures temporal dependencies through both forward and backward Mamba encoders, achieving high prediction accuracy. To reduce reliance on large datasets and lower training costs, TSMamba employs a two-stage transfer learning process that leverages pretrained Mamba LLMs, allowing effective time series modeling with a moderate training set. In the first stage, the forward and backward backbones are optimized via patch-wise autoregressive prediction; in the second stage, the model trains a prediction head and refines other components for long-term forecasting. While the backbone assumes channel independence to manage varying channel numbers across datasets, a channel-wise compressed attention module is introduced to capture cross-channel dependencies during fine-tuning on specific multivariate datasets. Experiments show that TSMamba’s zero-shot performance is comparable to state-of-the-art time series foundation models, despite using significantly less training data. It also achieves competitive or superior full-shot performance compared to task-specific prediction models. The code will be made publicly available.
		
	\end{abstract}

	\section{Introduction}
	
	Time series forecasting predicts future data based on historical chronological information, offering a valuable tool for anticipating changes, formulating strategies, and mitigating risks. This technique is widely used across various sectors, including energy, finance, healthcare, manufacturing, retail, and traffic management.
	
	Given the dynamic and ever-evolving nature of real-world data, forecasting models should be capable of adapting to changing patterns. However, traditional supervised models trained or even designed for each individual dataset or tasks (referred to as specialized models hereinafter), which are commonly used for time series forecasting, are often static and struggle to accommodate evolving patterns. This issue stems from three main challenges: first, these models require specific datasets for training, yet relevant data for emerging patterns may be unavailable or difficult to collect; second, they lack the ability to generalize across different datasets or applications, making it expensive and time-consuming to adapt models from one domain to another; and third, they often exhibit low data efficiency, increasing the risk of overfitting when training data are limited.
	
\begin{figure*}
		\centering
		\includegraphics[width=0.9 \linewidth]{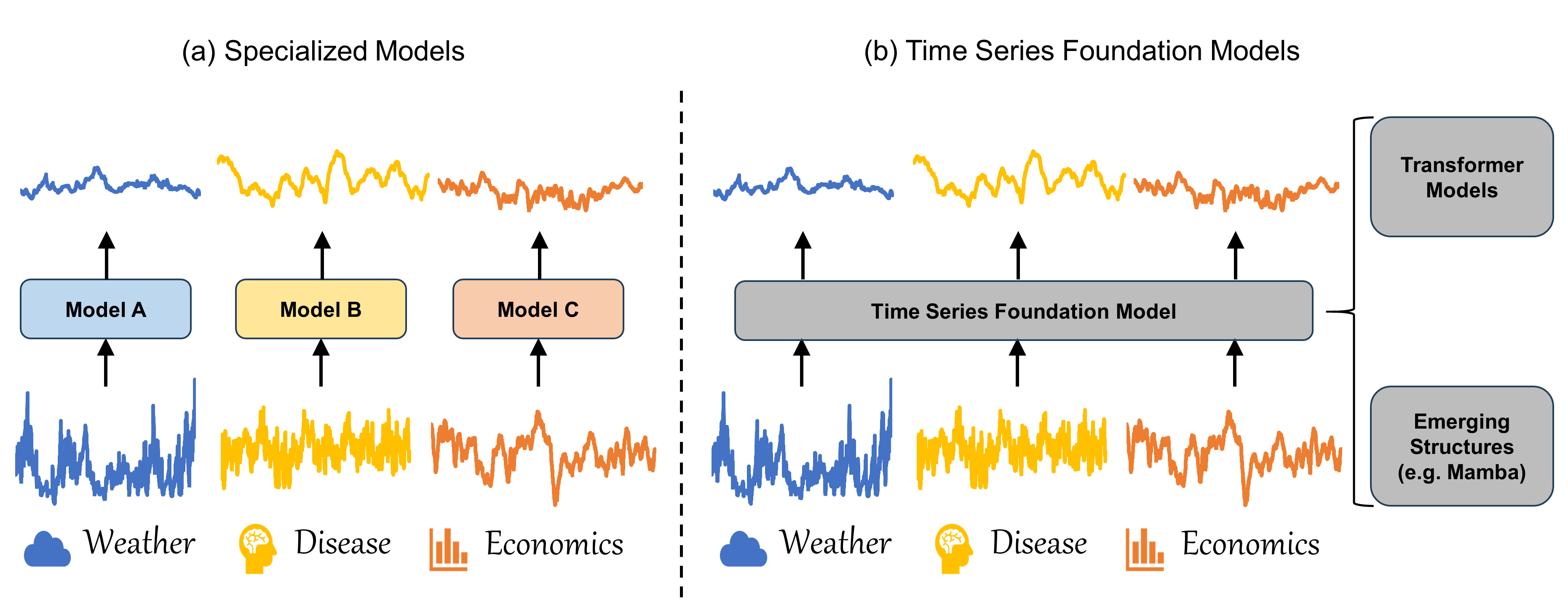}
		\caption{Comparison of specialized time series models and foundation models: (a) Specialized models are trained separately for specific tasks using relevant datasets. These models lack the ability to generalize across different domains and frequencies. (b) Time series foundation models, trained on large datasets, generalize well across a wide range of scenarios and tasks.}
		\label{FIG_SPEC_FM}
	\end{figure*}

	In contrast, time series foundation models, which are pretrained on vast domains of data, have demonstrated strong generalization capabilities across a wide range of scenarios and tasks. These models also exhibit high data efficiency in fine-tuning, enabling them to adapt to specific datasets with minimal samples. Such advantages make them effective for forecasting emerging patterns in web data, even when relevant data are unavailable or scarce. Comparison of specialized models and time series foundation models are presented in figure \ref{FIG_SPEC_FM}.
	
	The development of foundational models for time series forecasting draws inspiration from the success of large language models (LLMs, e.g., \citealt{devlin2018bert, brown2020gpt3, touvron2023llama}) in natural language processing (NLP), though it faces additional challenges.
 
	The first challenge is the significant heterogeneity of time series data. Data from different domains exhibit diverse dependencies, and data collected at varying frequencies or sampling rates present distinct patterns. Additionally, data gathered from different devices show varying noise levels. Multivariate time series also differ in their channel characteristics, with each dataset potentially containing a different number of channels. In contrast, text data for NLP models generally do not involve concepts like frequencies or sampling rates and typically consist of a single channel.
 
	Secondly, acquiring large-scale time series data is more challenging than in NLP. For example, the Large-scale Open Time Series Archive (LOTSA, \citealt{woo2024unified}), the largest publicly available time series dataset to our knowledge, contains around 27 billion time points, whereas large NLP datasets, such as RedPajama-Data-v2 \cite{together2023redpajama}, include tens of trillions of tokens.
	
	Lastly, training foundational models for time series on large datasets imposes enormous computational demands, leading to long training times and high resource consumption. As a result, the process is both time-intensive and costly, requiring a significant budget for computing power.
	
	Motivated by these challenges, some emerging time series foundation models (e.g., \citealt{zhou2023gpt4ts, jin2024timellm, liu2024calfaligningllmstime, liu2024unitime}) leverage existing LLMs and adapt them to time series tasks through cross-modality transfer learning. This approach allows these models to harness the knowledge acquired during LLM pretraining on vast datasets using significant computational resources. Other models are trained directly on large-scale time series datasets (e.g., \citealt{garza2023timegpt, das2023decoder, woo2024unified}), many of which are collected by the researchers themselves. However, several of the largest and most valuable datasets remain inaccessible to the public, and training with such large datasets also results in enormous computational costs.
	
	Existing time series foundation models are predominantly based on the Transformer architecture \cite{transformer}, which suffers from two main drawbacks: quadratic complexity with respect to input length and a lack of inductive biases \cite{d2021convit}, which are advantageous for leveraging the chronological order of data. Recently, structured state space sequence models (SSMs) \cite{gu2021, gu2022} have emerged as an efficient approach for sequence modeling, offering linear complexity. Mamba \cite{gu2023mamba} further enhances these SSMs by making the parameters functions of the inputs, allowing the model to selectively propagate or forget information along the sequence dimension based on the current data. Additionally, Mamba implements a hardware-aware parallel algorithm for efficient computation. The model achieves performance comparable to Transformers in NLP \cite{waleffe2024empirical} and CV \cite{zhu2024vision}, offering a strong alternative architecture for time series foundation models.
	
	This paper introduces TSMamba, a time series foundation model based on the Mamba architecture for multivariate forecasting. The model combines forward and backward Mamba encoders to capture temporal dependencies with linear complexity, achieving high predictive accuracy. To address the challenges of limited dataset sizes and training budgets, we propose a two-stage transfer learning process that leverages knowledge from large-scale pretraining of Mamba LLMs, allowing efficient adaptation to the time series modality. Additionally, while the pretrained model assumes channel independence (CI) to handle varying channel numbers, we introduce a compressed cross-channel attention module to capture cross-channel dependencies during fine-tuning on specific datasets.
	
	Our key contributions are as follows:
	
	\begin{itemize} 
		\item We propose TSMamba, a linear-complexity foundation model for time series forecasting, applicable to prediction tasks across different domains and frequencies. \item The two-stage transfer learning process allows the model to leverage relationships distilled from large-scale LLM pretraining, enabling effective adaptation to time series data while mitigating the need for large datasets and extensive training costs.
		\item We introduce a channel-wise compressed attention module that enables the model to extract cross-channel dependencies during fine-tuning, outperforming channel independent approaches in most datasets and settings. 
		\item The model achieves state-of-the-art (SOTA) performance on multiple mainstream datasets in zero-shot and full-shot forecasting scenarios. 
	\end{itemize}

	\section{Related Work}
	
	Over the last decade, time series forecasting has evolved from traditional statistical approaches to more advanced deep neural network-based techniques. Traditional models such as Autoregressive (AR), ARIMA, and VAR \citep{TimeSeriesAnalysisBook}, as well as kernel methods \citep{kernel-method} and Gaussian processes \citep{Gaussian-Process_0}, have been widely used. However, the advent of deep learning, fueled by the rapid evolution of computing capabilities and neural network architectures, has marked a paradigm shift in this field.
	
	Various deep neural network architectures, including recurrent neural networks (RNNs), convolutional neural networks (CNNs), graph neural networks (GNNs), multi-layer perceptrons (MLPs), and Transformers, have been extensively applied to time series forecasting. RNNs (e.g., \citealt{LSTM, Da-RNN, DeepSS, DeepAR}), specifically designed for sequential data, were among the first deep learning models utilized in this domain. However, RNNs encountered challenges such as vanishing gradients \cite{Pascanu2012OnTD}, which limited their effectiveness in capturing long-term dependencies. Similarly, CNNs, originally developed for image processing, were adapted to achieve state-of-the-art performance in time series forecasting \citep{tcn, Borovykh2017, Sen2019, micn, timesnet, dong2024moderntcn}. While CNNs excel at identifying local patterns, their limited receptive field size constrained their ability to capture long-term dependencies. GNNs are increasingly being employed to enhance the recognition of both temporal and dimensional patterns in time series data \citep{Wu2020, Cao2020}. Additionally, recent advancements \citep{nhits, Li2023mts, dlinear, tide, tsmixer} suggest that MLP-based architectures remain competitive in forecasting tasks.
	
	Following the remarkable success of Transformers \citep{transformer} in NLP \citep{nlpsurvey}, computer vision \citep{cvsurvey}, and speech processing \citep{speechsurvey}, Transformers have become a mainstream approach in time series forecasting, delivering promising results \citep{tssurvey}. Their attention mechanisms are highly effective at capturing long-term dependencies. However, Transformers suffer from quadratic complexity in both computation and memory. Additionally, their flexibility leads to a lack of certain inductive biases \cite{d2021convit}, which are beneficial for extracting sequential temporal dependencies.
	
	Various Transformer-based approaches have been proposed to simultaneously enhance forecasting performance and reduce computational costs. For example, Informer \citep{zhou2021informer} and Autoformer \citep{wu2021autoformer} reduce complexity to \(O(L\log(L))\), where \(L\) is the input length. Several improved Transformers \citep{linformer, luna, nystroformer, performer, pyraformer, fedformer, chen2024jtft} even achieve linear complexity. PatchTST \cite{patchtst} applies the patching technique \citep{vit, beit, mae} to the context of time series, dividing time series into overlapping or non-overlapping continuous patches and embedding each patch instead of individual time points. Although patching does not reduce the theoretical quadratic complexity, it substantially lowers the actual computational costs.
	
	The success of foundation models in NLP, which utilize large-scale pre-training to tackle diverse tasks with minimal labeled data, has inspired similar strategies in time series forecasting. Although pretraining requires extensive computational resources, these models can be fine-tuned and deployed for specific prediction tasks with moderate training budgets.
	
	To address the scarcity of large datasets for training time series foundation models, recent efforts have adapted pre-trained large language models (LLMs) to create a unified framework for various time series tasks, effectively leveraging the ability of transformer-based models to generalize across different domains. Among these efforts, \citet{zhou2023gpt4ts} demonstrated that the self-attention mechanism functions similarly to PCA, offering a deeper understanding of the universality of transformer-based models. They leveraged a primarily frozen GPT-2 backbone \cite{radford2019gpt2} to achieve competitive performance across a range of time series tasks. TIME-LLM \citep{jin2024timellm} converts input time series data into text prototype representations and enhances input context by incorporating declarative prompts to effectively guide the LLM's reasoning process. Additionally, \citet{chang2023llm4ts} developed a two-stage fine-tuning approach to adapt the GPT-2 backbone model for time series forecasting tasks. 
	
	On the other hand, some works \cite{garza2023timegpt,woo2024unified,liu2024timer,goswami2024moment} focus on collecting large datasets and training models directly on these datasets, achieving prominent zero-shot or few-shot performance across a variety of tasks, even closely matching the accuracy of supervised forecasting models tailored to each dataset.
	
	Almost all these time series foundation models are based on the Transformer architecture, which means they share the drawbacks of quadratic complexity and lack of inductive bias. 
	
	The emergence of state space models (SSMs, \citealt{gu2021, gu2022, wang2022pretraining, smith2023simplified}), offering linear or near-linear scaling and improved long-range dependency capture, has spurred significant advances in sequence modeling. By combining principles from RNNs and CNNs, SSMs enable efficient computation and excel in continuous signal domains like audio and vision \citep{goel2022s, saon2023diagonal}. However, a key weakness of these models is their inability to perform content-based reasoning. To address this issue, Mamba \citep{gu2023mamba} introduces a selective mechanism that efficiently filters and retains relevant information. It also presents a hardware-aware parallel algorithm, ensuring both theoretical linear complexity and improved practical computational efficiency. \citet{dao2024mamba2} further elucidate the theoretical connections between SSMs and variants of attention. The promising Mamba model has been applied in various scenarios (e.g., \citealt{zhu2024vision,waleffe2024empirical, wang2024graph}) and offers an alternative architecture for time series foundation models.

	\section{Method}
	
	This section applies the advanced state space model, Mamba, to construct a foundational model for time series forecasting, called TSMamba. We begin with an introduction to the time series forecasting problem, followed by a description of the Mamba model. Next, we outline the structure of the foundational model and propose a two-stage transfer learning approach to adapt the model to time series data across different domains and frequencies. Finally, we present the fine-tuning process designed to extract relationships within specific datasets, incorporating a compressed channel-wise attention module to leverage cross-channel dependencies.
	
	\subsection{The time series forecasting problem}
	
	We consider the multivariate time series forecasting problem, which involves predicting the future values of a time series based on historical data. The input series is denoted by \(\mathbf{x}_{1:L}=\{\mathbf{x}_1,\cdots,\mathbf{x}_L\}\), where \(L\) represents the look-back window (input length). The value at the \(i\)th time step is \(\mathbf{x}_i \in \mathcal{R}^D\), where \(D\) is the number of channels. The model maps the input to the prediction \(\mathbf{Y}\in \mathcal{R}^{D\times T}\), where \(T\) is the prediction length, also known as the target window. The goal of the model is to understand the input series and minimize the prediction error relative to the actual future values \(\mathbf{x}_{L+1:L+T}\).
	
	\subsection{Preliminaries of Mamba}
	
	Originating from the classic Kalman filter model \cite{kalman1960}, the state space model (SSM) has recently garnered significant interest. Among these methods, structured state space sequence models (S4) and Mamba represent a recent class of sequence models inspired by the following continuous system
	\begin{equation}
		\begin{aligned}
			&\mathbf{h}'(t)=\mathbf{A}\mathbf{h}(t) +\mathbf{B}x(t), \\
			&y(t)=\mathbf{C}\mathbf{h}(t).
			\label{SSM_CONTINUOUS}
		\end{aligned}
	\end{equation}
	This system maps a one-dimensional function or sequence $\mathbf{x}(t)\in \mathcal{R}$ to $\mathbf{y}(t)\in \mathcal{R}$ through an implicit latent state $\mathbf{h}(t)\in \mathcal{R}^{N_\text{st}}$.
	
	The system (\ref{SSM_CONTINUOUS}) can be discretized through a zero-order hold (ZOH) rule into
	\begin{equation}
		\begin{aligned}
			&\mathbf{h}_t=\mathbf{\bar{A}}\mathbf{h}_{t-1} +\mathbf{\bar{B}}x_t, \\
			&y_t=\mathbf{\bar{C}}\mathbf{h}_t,
			\label{SSM_DISC}
		\end{aligned}
	\end{equation}
	where
	\begin{equation}
		\begin{aligned}
			&\mathbf{\bar{A}}=\exp(\Delta\mathbf{A}), \\
			&\mathbf{\bar{B}},=(\Delta\mathbf{A})^{-1}\exp(\Delta\mathbf{A}-\mathbf{I})\cdot \Delta \mathbf{B}
			\label{COE_DISC}
		\end{aligned}
	\end{equation}
	and $\Delta$ is the time step.
	
	Mamba improves upon S4 by making the parameters that affect interactions along the sequence input-dependent, enabling the model to selectively propagate or forget information based on the current inputs, thereby enhancing its capability to perform content-based reasoning. 
	
	Specifically, it sets 
	\begin{equation}
		\begin{aligned}
			&\mathbf{B}=\text{Linear}_{N_\text{st}} (x), \\
			&\mathbf{C}=\text{Linear}_{N_\text{st}} (x), \\
			&\Delta=\text{softplus}(\text{Parameter}+\text{Broadcast}_{D_{\text{mb}}} (\text{Linear}_1(x)),
			\label{MAMBA_PARA_MAP}
		\end{aligned}
	\end{equation}
	where $\text{Linear}_d$ is a parameterized projection to dimension $d$, and $D_{\text{mb}}$ is the model dimension.
	
	\citet{gu2023mamba} also designed the Mamba block, which integrates an SSM into the main branch of the Gated MLP block and serves as a foundational module for building LLMs of different scales.
	
	The modifications in Mamba introduce some drawbacks in parallel computation. S4 can be computed in two ways: either by using the linear recurrence form (\ref{SSM_DISC}), which is efficient for autoregressive inference, or by employing a global convolution mode, which can be parallelized effectively during training \cite{gu2022}. However, when Mamba introduces input-dependent parameters, it becomes incompatible with the convolution mode, making it difficult to fully utilize the strong parallel processing power of hardware accelerators such as GPUs. 
	
	To address this, \citet{gu2023mamba} designed a hardware-aware parallel algorithm in recurrent mode. \citet{dao2024mamba2} further improves the Mamba model, enabling the implementation of tensor parallelism. Consequently, Mamba not only achieves theoretically linear scaling with sequence length but also benefits from fast training and inference in practice.

	\subsection{Architecture of the Foundation Model}
	
	\begin{figure*}[t]
		\centering
		\includegraphics[width=0.9 \linewidth]{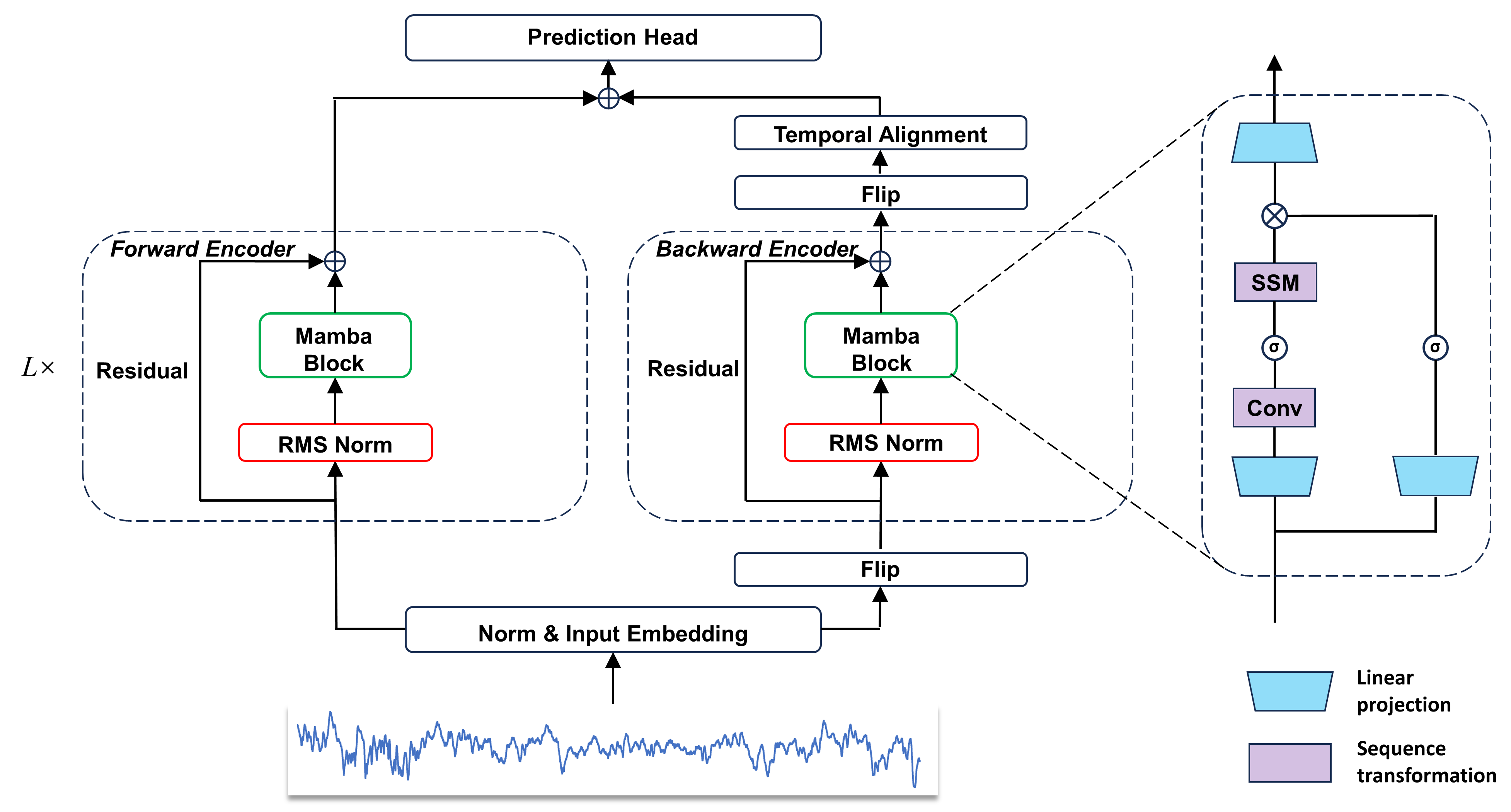}
		\caption{TSMamba Architecture: The input time series are preprocessed and then fed into the forward and backward encoder to extract internal dependencies. The representations are combined and subsequently mapped to forecasts by the prediction head.}
		\label{FIG_ARCH}
	\end{figure*}
	
	The architecture of TSMamba is illustrated in Figure \ref{FIG_ARCH}. The model encodes preprocessed data using a backbone comprised of forward and backward Mamba encoders. These encoders consist of homogeneously stacked Mamba blocks, interspersed with standard normalization and residual connections. While the forward encoder extracts sequential causal dependencies, the backward encoder enriches the representation by capturing inverse time relations from the flipped embedding. The backward representation is then flipped and processed through a temporal convolution module to align with the forward representation in the time dimension. Finally, the combined representations are mapped to the forecasting output by a prediction head.
	
	The preprocessing module consists of normalization and input embedding. Given that the number of channels varies across datasets, the multivariate time series are processed in a channel-independent (CI) setting, where each variate is treated as a univariate series. Since normalization is crucial for effective knowledge extraction, we incorporate reverse instance normalization \cite{kim2021reversible}, which normalizes the input time series using mean and variance. 
	
	The input embedding is implemented using a 1D convolution layer, which functions similarly to patching \cite{patchtst} but offers greater convenience. In the patching approach, each univariate series is divided into either overlapping or non-overlapping continuous segments, each of which is then embedded into a vector via linear mapping. This technique preserves local semantic information within the embedding, enhancing the model's ability to capture comprehensive semantic details that might be overlooked at the individual point level. Additionally, it reduces the input length for encoders. We apply a 1D convolution to provide a simplified implementation of patching. In this implementation, the input channel is set to 1, the output channel is set to the model dimension $D_m$, and the stride is set to the patch length $p_l$. This module effectively maps non-overlapping segments of $p_l$ time points to a vector embedding of dimension $D_m$.
	
	TSMamba employs both a forward and a backward encoder (referred to as the backbone hereinafter) to extract temporal dependencies. The forward encoder captures sequential causal relations, while the backward encoder provides additional information by leveraging inverse time relations. The output of the backward encoder is flipped and aligned with the forward representation using a convolution along the time dimension.

	Both the forward and backward encoders leverage the backbone of Mamba language models, where each layer consists of a Mamba block equipped with RMSNorm and a residual connection. The Mamba block integrates a state space model (SSM) into the main branch of the Gated MLP block. The block expands the input model dimension \(D_m\) to the inner dimension \(D_{\text{mb}}\) by a factor of 2, then contracts it back to \(D_m\), allowing for homogeneous stacking of blocks. The activation function used is SiLU \cite{elfwing2018silu}. Unlike the commonly used Transformer encoder or decoder blocks, which require a KV cache proportional to the number of historical tokens, the Mamba block propagates only a fixed-size internal state along the time dimension. This design enables linear complexity in both time and space.
	
	The prediction head generates forecasts based on the historical data representations extracted by the two encoders. In a canonical Transformer decoder, predictions are made autoregressively, where a linear head maps the representation of the last token to outputs, which are then added to the inputs to predict subsequent tokens. However, \citet{patchtst} show that using a larger linear head to map the representations of all historical patches to the entire target window at once can improve predictions by reducing error accumulation compared to the autoregressive approach.
	
	In a foundation model, where the model dimension is significantly larger than in specialized models, using a linear head that considers all historical representations would result in an excessive number of parameters, making the model prone to overfitting. To address this, the prediction head first compresses the model dimension using a linear projection with GELU \cite{gelu} activation. It then maps the compressed representations to the target window with a much smaller linear head. Finally, the outputs are denormalized to restore the original mean and variance.
	
	The model is trained with the Huber loss function, which offers enhanced robustness to outliers in the data compared to the Mean Squared Error (MSE) loss.
	
	\subsection{Two-stage transfer learning approach}
	
	To harness the knowledge of existing language models, enabling TSMamba to adapt to time series data across different domains and frequencies with low training costs, we designed a two-stage transfer learning approach for training the model.
	
	\begin{figure*}
		\centering
		\includegraphics[width=0.9\linewidth]{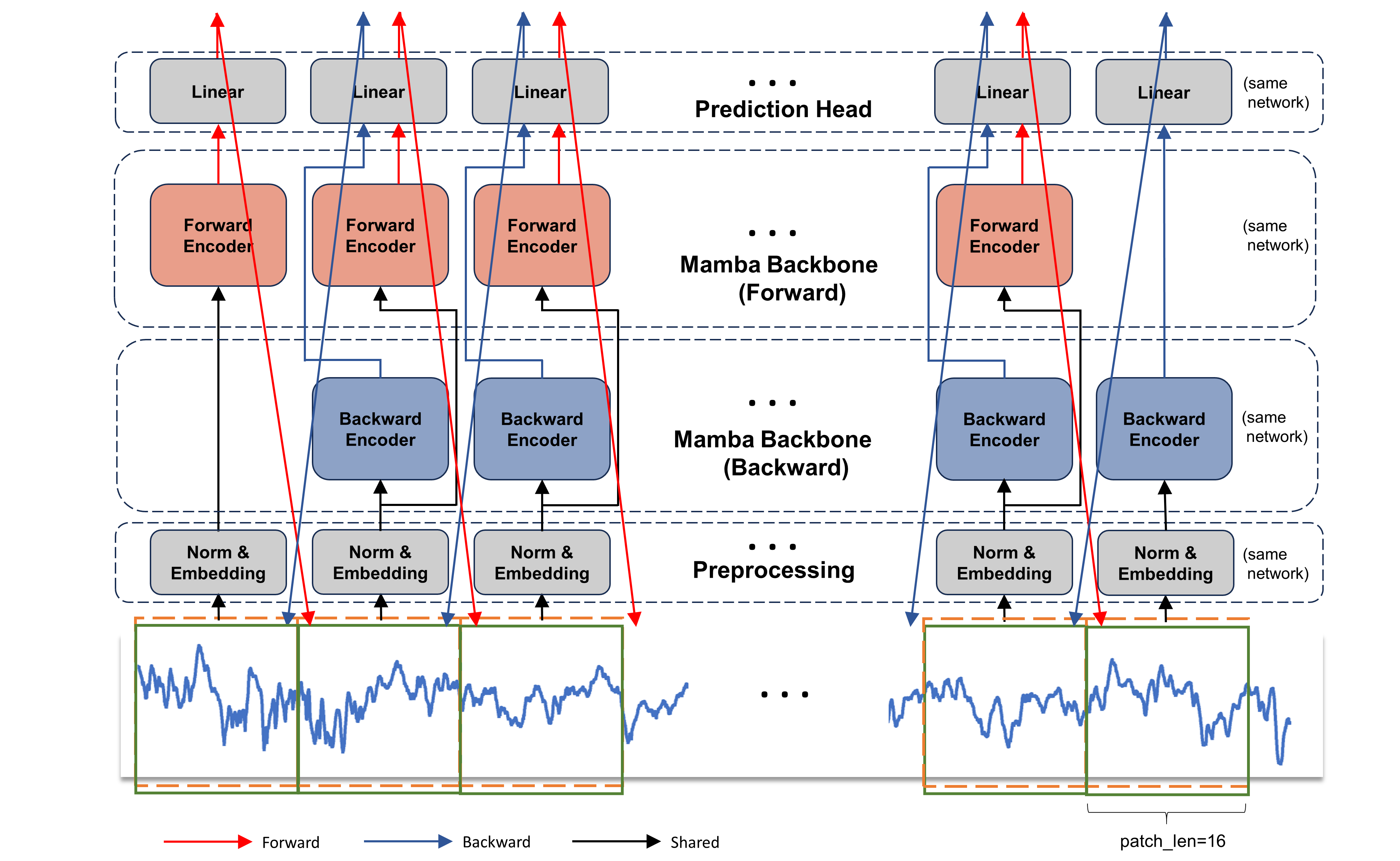}
		\caption{The first stage of transfer learning involves refining the backbone and training the input embedding through autoregressive forecasting or backcasting tasks. A small linear head is temporarily added to predict the next patch.}
		\label{FIG_AUTOREGRESSIVE}
	\end{figure*}
	
	The first stage involves refining the backbone and training the input embedding through autoregressive forecasting or backcasting tasks. 
	
	As shown in Figure \ref{FIG_AUTOREGRESSIVE}, the model architecture undergoes slight modifications during this stage. The prediction head, which originally handled compressed historical patches, is replaced with a smaller head that uses the representation of the current patch to predict the next one with the forward representation and the last one with the backward representation. Although a merely autoregressive forecasting is not ideal for long-term forecasting due to error accumulation, this stage is crucial for refining the representation of each patch. This is because the supervision signal for predicting the next patch or the last patch is focused on the current patch's representation, rather than being dispersed across all historical patches as in the original large head configuration. 
	
	The backbone is initialized with the Mamba language model, specifically Mamba-130M, while the input embedding and prediction head are trained from scratch.
	
	The second stage focuses on training the prediction head and further refining the other structures. The original TSMamba architecture is restored, with the backbone and input embedding loaded from the results of the first stage, while the prediction head is randomly initialized. This stage produces the TSMamba foundation model for forecasting, which can be used for zero-shot predictions directly or fine-tuned to further enhance performance on specific datasets.
	
	In this stage, the newly initialized prediction head and temporal alignment module are trained with a larger learning rate, while the existing backbone and embedding are updated with a smaller learning rate to fully leverage the pretrained model from the first stage.
	
	The two-stage transfer learning approach also extends to downstream tasks beyond forecasting, such as imputation, classification, and anomaly detection. In these cases, the first stage provides an input embedding and a refined backbone that produce a robust representation of the time series data. The second stage can then be adapted to accommodate specific tasks, achieving zero-shot performance using the information from the training sets.
	
	\subsection{Fine-tuning with cross-channel relation extraction}
	
	Fine-tuning focuses on learning the unique dependencies of a specified dataset, typically with limited training data. During fine-tuning, we freeze the Mamba blocks in the backbone, which contain most of the model's parameters, to preserve the relationships extracted during pretraining and the two-stage transfer learning. The input embedding, RMSNorm, and prediction head are adjusted to learn from the new dataset. Additionally, we introduce a compressed channel-wise Transformer encoder to extract cross-channel relations.
	
	The compressed channel-wise attention module is added before the prediction head to extract cross-channel dependencies. 
	
	Although leveraging cross-channel information is intuitive, previous works \cite{patchtst, chen2024jtft} have shown that it is challenging to utilize these relationships effectively to improve predictions, given the strong baseline established by channel-independent (CI) models. The difficulty arises because typical time series datasets are not large enough to disentangle time-wise and channel-wise modeling effectively. CI models, which treat each variate as an independent sequence, provide more samples of univariate series, thereby reducing the risk of overfitting.
	
	\begin{figure}
		\centering
		\includegraphics[width=0.5\linewidth]{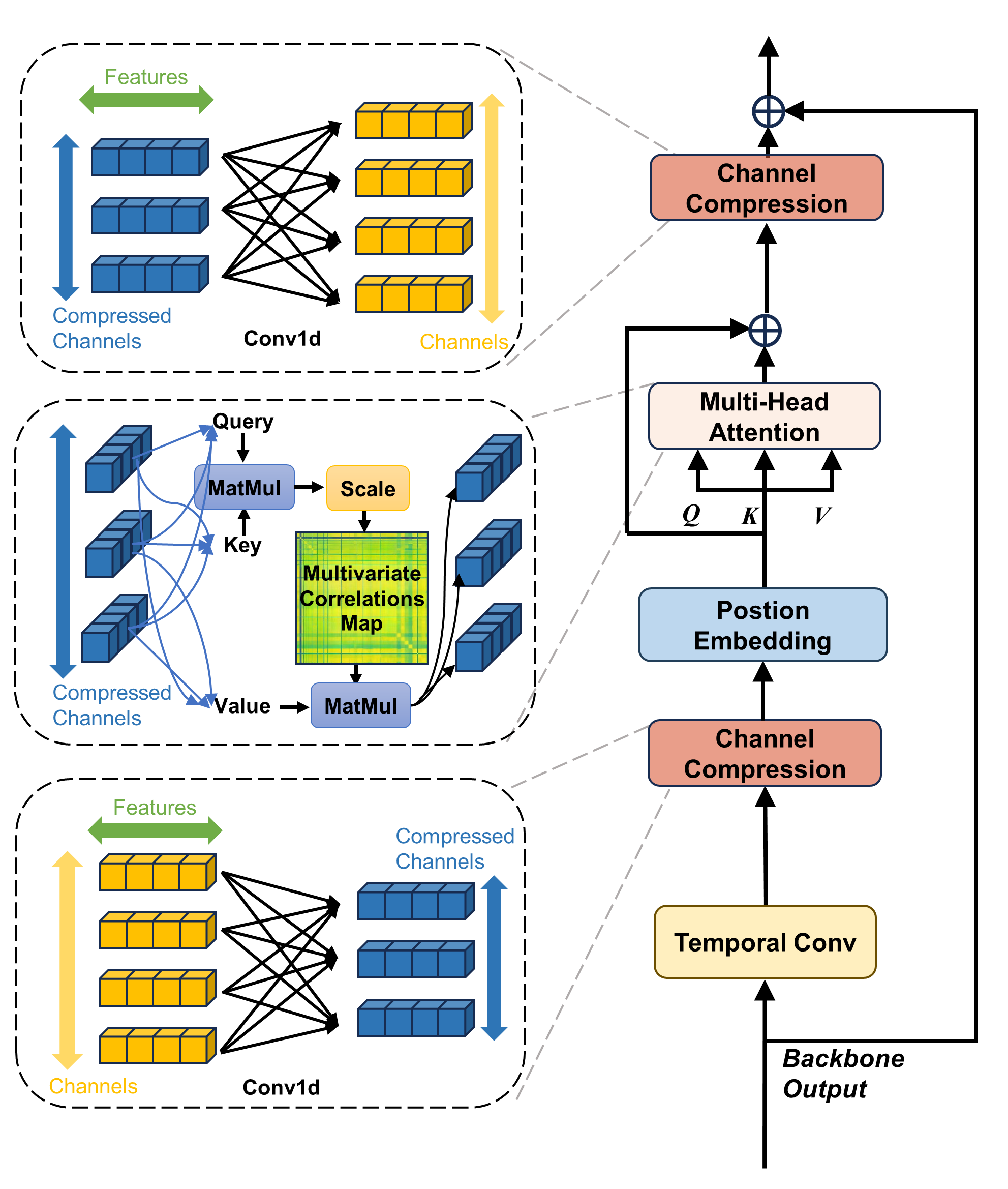}
		\caption{Compressed Channel-Wise Attention Module for Cross-Channel Dependency. The process starts with a per-channel temporal convolution to align the backbone outputs along the time dimension, followed by linear compression of the channel count. The attention module then extracts relationships between these compressed channels, and the result is linearly mapped to restore the original number of channels. Finally, the output is added back to the backbone as a correction.}
		\label{FIG_CROSS_CH}
	\end{figure}
	
	As experiments have shown that linear mixing of channels or adding a channel-wise Transformer encoder often degrades performance, we chose to apply an attention module on a compressed channel dimension, as illustrated in Figure \ref{FIG_CROSS_CH}. The module takes the representation extracted by the backbone as input. First, we apply a temporal convolution for each channel, acting as a time shift to align channels, since the most relevant data across different channels may exhibit time lags. Next, the number of channels is reduced from \(D\) to \(\lceil\log_2(D)\rceil\) via linear projection, where \(\lceil\cdot\rceil\) denotes the ceiling operation. The attention module is then applied to extract dependencies along the compressed channel dimension. The output is subsequently linearly mapped back to restore the \(D\) channels and is added as a correction to the backbone representation. This channel compression acts as a regularization mechanism, filtering out noise in cross-channel relations while reducing computational costs, particularly when \(D\) is large. In practice, the linear compression and expansion of channels are performed using 1D convolutions with a kernel size of 1, preventing any permutations.
	
	Since the process of extracting cross-channel dependencies is prone to overfitting, the module is activated only when sufficient training data are available.
	
	\section{Experiments}
	
	We compare TSMamba with 16 different baselines, which represents state-of-the-art models in long-term forecasting. Our baselines could be divided into two parts: zero-shot forecasting and full-shot forecasting. The zero-shot forecasting includes many pretrained foundation time-series models such as Time-MoE \citep{shi2024time}, Moirai \citep{woo2024unified}, TimeFM \citep{das2023decoder}, Timer \citep{liu2024timer}, Moment \citep{goswami2024moment}, and Chronos \citep{ansari2024chronos}. As for the full-shot forecasting evaluation includes a collection of Transformer-based models such as PatchTST \citep{patchtst}, Autoformer \citep{wu2021autoformer}, FEDformer \citep{fedformer}, and Crossformer \citep{zhang2023crossformer}.  Additionally, we incorporate large language models (LLMs) like GPT4TS \citep{zhou2023gpt4ts} and CALF \citep{liu2024calfaligningllmstime}, as well as other recent competitive approaches, including LightTS \citep{zhang2022less}, DLinear \citep{dlinear}, RLinear \citep{li2023revisiting}, and MambaTS \citep{cai2024mambats}. 

    The configuration of TSMamba includes 3 encoder layers, an embedding size of 768, and a fixed input sequence length of 512.

    \subsection{ZERO-SHOT FORECASTING}

    The zero-shot forecasting results assess the generalizability of foundation models by examining each model’s ability to adapt to previously unseen data.
    
    The models are evaluated on widely recognized long-term forecasting benchmarks that differ from those used during pre-training. As this work is in progress, TSMamba is evaluated on ETTm2 and Weather, two commonly used medium-sized datasets. For each dataset, we considered four different forecasting horizons: \{96, 192, 336, 720\}. Model performance is measured using two standard evaluation metrics: mean squared error (MSE) and mean absolute error (MAE). The results of other methods for comparison are obtained from \citep{shi2024time}.
    
    Results. Our brief zero-shot results are shown in table \ref{tab:zero_shot_full}, comparing TSMamba with state-of-the-art models. While TSMamba does not consistently outperform all baselines, it excels at longer prediction lengths (336 and 720) and achieves competitive average performance. Notably, despite being pre-trained on significantly less data, TSMamba performs comparably to models with larger pre-training datasets, demonstrating its data efficiency and robustness in zero-shot forecasting. 

     \vspace{-3mm}
\begin{table}[htbp]
  \centering
  \caption{Full results of zero-shot forecasting experiments. TimesFM, pre-trained on Weather datasets, is excluded from evaluation on these datasets, and its results are represented by a dash ($-$). {\boldres{Red}}: the best, \secondres{Blue}: the 2nd best.} 
  \vspace{5mm}
  \resizebox{\columnwidth}{!}{
    \renewcommand{\tabcolsep}{3pt}
    \begin{tabular}{cr|cc|cc|cc|cc|cc|cc|cc|cc|cc|cc|cc|cc}
          \toprule
          \multicolumn{2}{c}{\multirow{3}{*}{\textbf{\scalebox{1.2}{Models}}}} & \multicolumn{20}{c}{\textbf{Zero-shot Time Series Models}} \\
          \cmidrule(lr){3-24}
          &       &\multicolumn{2}{c}{\textbf{{TSMamba}}}& \multicolumn{2}{c}{\textbf{{Time-MoE$_{base}$}}} & \multicolumn{2}{c}{\textbf{{Time-MoE$_{large}$}}} & \multicolumn{2}{c}{\textbf{Moirai$_{small}$}} & \multicolumn{2}{c}{\textbf{Moirai$_{base}$}} & \multicolumn{2}{c}{\textbf{Moirai$_{large}$}} & \multicolumn{2}{c}{\textbf{TimesFM}} & \multicolumn{2}{c}{\textbf{Moment}} & \multicolumn{2}{c}{\textbf{Chronos$_{small}$}} & \multicolumn{2}{c}{\textbf{Chronos$_{base}$}} & \multicolumn{2}{c}{\textbf{Chronos$_{large}$}} \\
          \cmidrule(lr){3-4} \cmidrule(lr){5-6}\cmidrule(lr){7-8} \cmidrule(lr){9-10}\cmidrule(lr){11-12}\cmidrule(lr){13-14}\cmidrule(lr){15-16}\cmidrule(lr){17-18}\cmidrule(lr){19-20}\cmidrule(lr){21-22}\cmidrule(lr){23-24}
          
          \multicolumn{2}{c}{\scalebox{1.2}{\textbf{Metrics}}}& \textbf{MSE} & \textbf{MAE} & \textbf{MSE} & \textbf{MAE} & \textbf{MSE} & \textbf{MAE} & \textbf{MSE} & \textbf{MAE} & \textbf{MSE} & \textbf{MAE} & \textbf{MSE} & \textbf{MAE} & \textbf{MSE} & \textbf{MAE} & \textbf{MSE} & \textbf{MAE} & \textbf{MSE} & \textbf{MAE} & \textbf{MSE} & \textbf{MAE} & \textbf{MSE} & \textbf{MAE}  \\
          \midrule
          \multirow{4}[0]{*}{ETTm2} & 96    & 0.201 & 0.287 & 0.201 & 0.291 & \boldres{0.197} & 0.286 & 0.214 & 0.288 & 0.205 & 0.273 & 0.211 & 0.274 & 0.202 & \boldres{0.270} & 0.260 & 0.335 & 0.209 & 0.291 & \secondres{0.199} & 0.274 & \boldres{0.197} & \secondres{0.271} \\
          & 192   & 0.259 & 0.325 & 0.258 & 0.334 & \boldres{0.250} & 0.322 & 0.284 & 0.332 & 0.275 & \secondres{0.316} & 0.281 & 0.318 & 0.289 & 0.321 & 0.289 & 0.350 & 0.280 & 0.341 & 0.261 & 0.322 & \secondres{0.254} & \boldres{0.314} \\
          & 336   & \secondres{0.315} & 0.361 & 0.324 & 0.373 & 0.337 & 0.375 & 0.331 & 0.362 & 0.329 & \boldres{0.350} & 0.341 & 0.355 & 0.360 & 0.366 & 0.324 & 0.369 & 0.354 & 0.390 & 0.326 & 0.366 & \boldres{0.313} & \secondres{0.353} \\
          & 720   & 0.406 & 0.417 & 0.488 & 0.464 & 0.480 & 0.461 & \secondres{0.402} & \boldres{0.408} & 0.437 & 0.411 & 0.485 & 0.428 & 0.462 & 0.430 & \boldres{0.394} & \secondres{0.409} & 0.553 & 0.499 & 0.455 & 0.439 & 0.416 & 0.415 \\
          \rowcolor{tabhighlight}
          & {\textbf{AVG}} & \boldres{0.295} & 0.347 & 0.317 & 0.365  & 0.316 & 0.361 & 0.307 & 0.347 & 0.311 & \boldres{0.337} & 0.329 & 0.343 & 0.328 & 0.346 & 0.316 & 0.365 & 0.349 & 0.380 & 0.310 & 0.350 & \boldres{0.295} & \secondres{0.338} \\
    \midrule
    \multirow{4}[0]{*}{Weather} & 96    & 0.179 & 0.235 & \secondres{0.160} & 0.214 & \boldres{0.159} & \secondres{0.213} & 0.198 & 0.222 & 0.220 & 0.217 & 0.199 & \boldres{0.211} & - & - & 0.243 & 0.255 & 0.211 & 0.243 & 0.203 & 0.238 & 0.194 & 0.235 \\
          & 192   & 0.227 & 0.278 & \boldres{0.210} & 0.260 & \secondres{0.215} & 0.266 & 0.247 & 0.265 & 0.271 & \secondres{0.259} & 0.246 & \boldres{0.251} & - & - & 0.278 & 0.329 & 0.263 & 0.294 & 0.256 & 0.290 & 0.249 & 0.285 \\
          & 336   & \secondres{0.278} & 0.315 & \boldres{0.274} & 0.309 & 0.291 & 0.322 & 0.283 & 0.303 & 0.286 & \secondres{0.297} & \boldres{0.274} & \secondres{0.291} & - & - & 0.306 & 0.346 & 0.321 & 0.339 & 0.314 & 0.336 & 0.302 & 0.327 \\
          & 720   & \secondres{0.342} & 0.358 & 0.418 & 0.405 & 0.415 & 0.400 & 0.373 & 0.354 & 0.373 & \secondres{0.354} & \boldres{0.337} & \boldres{0.340} & - & - & \secondres{0.350} & 0.374 & 0.404 & 0.397 & 0.397 & 0.396 & 0.372 & 0.378 \\
          \rowcolor{tabhighlight}
          & {\textbf{AVG}} & \boldres{0.256} & 0.296 &  0.265 & 0.297 & 0.270 & 0.300 & 0.275 & 0.286 & 0.287 & \secondres{0.281} & \secondres{0.264} & \boldres{0.273} & - & - & 0.294 & 0.326 & 0.300 & 0.318 & 0.292 & 0.315 & 0.279 & 0.306 \\
    \midrule
    \rowcolor{blue!15}
    \multicolumn{2}{c|}{\scalebox{1.1}
    {\textbf{Average}}} & \boldres{0.275} & 0.321 & \secondres{0.291} & 0.331 & 0.293 & 0.331 & \secondres{0.291} & 0.317 & 0.299 & \secondres{0.309} & 0.297 & \boldres{0.308} & 0.328 & 0.346 & 0.305 & 0.346 & 0.325 & 0.349 & 0.301 & 0.333 & 0.287 & 0.322 \\
    \midrule
    \end{tabular}
    }
  \label{tab:zero_shot_full}%
  \vspace{-2mm}
\end{table}%

    \subsection{Full-SHOT FORECASTING}

    \begin{table}[htp]
  \centering
  \caption{Full results of in-domain forecasting experiments. A lower MSE or MAE indicates a better prediction. 
  {\boldres{Red}}: the best, \secondres{Blue}: the 2nd best.}
  
  \resizebox{\columnwidth}{!}{
    \renewcommand{\tabcolsep}{3pt}
    \begin{tabular}{cr|cc|cc|cc|cc|cc|cc|cc|cc|cc|cc|cc|cc}
          \toprule
            \multicolumn{2}{c}{\multirow{3}{*}{\scalebox{1.2}{\textbf{Models}}}}& \multicolumn{24}{c}{\textbf{Full-shot Time Series Models}} \\
          \cmidrule(lr){3-26}
          &       & \multicolumn{2}{c}{\textbf{{TSMamba}}}& \multicolumn{2}{c}{\textbf{CALF}} & \multicolumn{2}{c}{\textbf{PatchTST}} & \multicolumn{2}{c}{\textbf{MambaTS}} & \multicolumn{2}{c}{\textbf{GPT4TS}} & \multicolumn{2}{c}{\textbf{Autoformer}} &  \multicolumn{2}{c}{\textbf{FEDformer}} & \multicolumn{2}{c}{\textbf{LightTS}} & \multicolumn{2}{c}{\textbf{Crossformer}} & \multicolumn{2}{c}{\textbf{DLinear}} & \multicolumn{2}{c}{\textbf{TimesNet}} & \multicolumn{2}{c}{\textbf{RLinear}} \\

          \cmidrule(lr){3-4} \cmidrule(lr){5-6}\cmidrule(lr){7-8} \cmidrule(lr){9-10}\cmidrule(lr){11-12}\cmidrule(lr){13-14}\cmidrule(lr){15-16}\cmidrule(lr){17-18}\cmidrule(lr){19-20} \cmidrule(lr){21-22} \cmidrule(lr){23-24} \cmidrule(lr){25-26}
          
          \multicolumn{2}{c}{\scalebox{1.2}{\textbf{Metrics}}}& \textbf{MSE} & \textbf{MAE} & \textbf{MSE} & \textbf{MAE} & \textbf{MSE} & \textbf{MAE} & \textbf{MSE} & \textbf{MAE} & \textbf{MSE} & \textbf{MAE} & \textbf{MSE} & \textbf{MAE} & \textbf{MSE} & \textbf{MAE} & \textbf{MSE} & \textbf{MAE} & \textbf{MSE} & \textbf{MAE} & \textbf{MSE} & \textbf{MAE} & \textbf{MSE} & \textbf{MAE} & \textbf{MSE} & \textbf{MAE} \\
          \midrule
    \multirow{4}[0]{*}{ETTm2} & 96    & 0.168 & \secondres{0.256} & 0.178 & \secondres{0.256} & \secondres{0.166} & \secondres{0.256} & 0.174 & 0.269 & 0.173 & 0.262 & 0.255 & 0.339 & 0.180 & 0.271 & 0.209 & 0.308 & 0.421 & 0.461 & 0.167 & 0.260 & 0.187 & 0.267 & \boldres{0.164} & \boldres{0.253} \\
          & 192   & \secondres{0.222} & \secondres{0.292} & 0.242 & 0.297 & 0.223 & 0.296 & 0.235 & 0.309 & 0.229 & 0.301   & 0.281 & 0.340 & 0.252 & 0.318 & 0.311 & 0.382 & 0.503 & 0.519   & 0.224 & 0.303 & 0.249 & 0.309 & \boldres{0.219} & \boldres{0.290} \\
          & 336   & 0.279 & 0.335 & 0.307 & 0.339 & \secondres{0.274} & \secondres{0.329} & 0.288 & 0.346 & 0.286 & 0.341   & 0.339 & 0.372 & 0.324 & 0.364 & 0.442 & 0.466 & 0.611 & 0.580   & 0.281 & 0.342 & 0.321 & 0.351 & \boldres{0.273} & \boldres{0.326} \\
          & 720   & \boldres{0.357} & \boldres{0.383} & 0.397 & 0.393 & 0.362 & \secondres{0.385} & \secondres{0.360} & 0.393 & 0.378 & 0.401   & 0.433 & 0.432 & 0.410 & 0.420 & 0.675 & 0.587 & 0.996 & 0.750 & 0.397 & 0.421 & 0.497 & 0.403 & 0.366 & \secondres{0.385} \\
          \rowcolor{tabhighlight}
          & {\textbf{AVG}} & \secondres{0.257} & \secondres{0.317} & 0.281 & 0.321 & \boldres{0.256} & \secondres{0.317} & 0.264 & 0.329 & 0.266 & 0.326 & 0.327 & 0.371 & 0.292 & 0.343 & 0.409 & 0.436 & 0.632 & 0.578 & 0.267 & 0.332 & 0.291 & 0.333 & \boldres{0.256} & \boldres{0.314}  \\                      
    \midrule
    \multirow{4}[0]{*}{Weather} & 96    & \boldres{0.144} & \boldres{0.193} & 0.164 & 0.204 & 0.149 & 0.198 & \secondres{0.145} & \secondres{0.196} & 0.162 & 0.212 & 0.266 & 0.336 & 0.238 & 0.314 & 0.182 & 0.242 & 0.153 & 0.217 & 0.152 & 0.237 & 0.172 & 0.220 & 0.175 & 0.225 \\
          & 192   & \boldres{0.192} & \boldres{0.236} & 0.214 & 0.250 & 0.194 & \secondres{0.241} & \secondres{0.193} & \secondres{0.241} & 0.204 & 0.248 & 0.307 & 0.367 & 0.275 & 0.329 & 0.227 & 0.287 & 0.197 & 0.269 & 0.220 & 0.282 & 0.219 & 0.261  & 0.218 & 0.260 \\
          & 336   & \boldres{0.242} & \boldres{0.276} & 0.269 & 0.291 & \secondres{0.245} & \secondres{0.282} & 0.246 & 0.283 & 0.254 & 0.286 & 0.359 & 0.395 & 0.339 & 0.377 & 0.282 & 0.334 & 0.252 & 0.311 & 0.265 & 0.319 & 0.280 & 0.306 & 0.265 & 0.294 \\
          & 720   & \boldres{0.313} & \boldres{0.328} & 0.355 & 0.352 & \secondres{0.314} & 0.334 & \secondres{0.314} & \secondres{0.331} & 0.326 & 0.337 & 0.419 & 0.428 & 0.389 & 0.409 & 0.352 & 0.386 & 0.318 & 0.363 & 0.323 & 0.362 & 0.365 & 0.359 & 0.329 & 0.339 \\
          \rowcolor{tabhighlight}
          & {\textbf{AVG}} & \boldres{0.222} & \boldres{0.258} & 0.250 & 0.274 & 0.226 & 0.264 & \secondres{0.225} & \secondres{0.263} & 0.237 & 0.270 & 0.338 & 0.382 & 0.310 & 0.357 & 0.261 & 0.312 & 0.230 & 0.290 & 0.240 & 0.300 & 0.259 & 0.287 & 0.247 & 0.279  \\   
    \midrule
    \multirow{4}[0]{*}{ILI} & 96    & \boldres{1.189} & \boldres{0.659} & - & - & \secondres{1.319} & \secondres{0.754} & - & - & 2.063 & 0.881 & 2.906 & 1.182 & 2.624 & 1.095 & 8.313 & 2.144 & 3.040 & 1.186 & 2.215 & 1.081 & 2.317 & 0.934 & 4.337 & 1.507 \\
          & 192   & \boldres{1.227} & \boldres{0.687} & - & - & \secondres{1.430} & \secondres{0.834} & - & - & 1.868 & 0.892 & 2.585 & 1.038 & 2.516 & 1.021 & 6.631 & 1.902 & 3.356 & 1.230 & 1.963 & 0.963 & 1.972 & 0.920  & 4.205 & 1.481 \\
          & 336   & \boldres{1.246} & \boldres{0.700} & - & - & \secondres{1.553} & \secondres{0.815} & - & - & 1.790 & 0.884 & 3.024 & 1.145 & 2.505 & 1.041 & 7.299 & 1.982 & 3.441 & 1.223 & 2.130 & 1.024 & 2.238 & 0.940 & 4.257 & 1.484 \\
          & 720   & \boldres{1.357} & \boldres{0.762} & - & - & \secondres{1.470} & \secondres{0.788} & - & - & 1.979 & 0.957 & 2.761 & 1.114 & 2.742 & 1.122 & 7.283 & 1.985 & 3.608 & 1.302 & 2.368 & 1.096 & 2.027 & 0.928 & 4.278 & 1.487 \\
          \rowcolor{tabhighlight}
          & {\textbf{AVG}} & \boldres{1.255} & \boldres{0.702} & - & - & \secondres{1.443} & \secondres{0.798} & - & - & 1.925 & 0.903 & 2.819 & 1.120 & 2.597 & 1.070 & 7.382 & 2.003 & 3.361 & 1.235 & 2.169 & 1.041 & 2.139 & 0.931 & 4.269 & 1.490  \\   
    \midrule
    \end{tabular}%
    }
  \label{tab:lsf_full}%
\end{table}%

    To assess TSMamba’s adaptability to specific datasets through fine-tuning, we compare its forecasting results on three widely used datasets: ILI, ETTm2, and Weather. TSMamba is tested with four different prediction horizons \(\{96, 192, 336, 720\}\). The results of other methods for comparison are obtained from \citep{patchtst, zhou2023gpt4ts, cai2024mambats}.

    Results. Our brief full-shot results are shown in Table \ref{tab:lsf_full}, where TSMamba demonstrates superior performance across multiple datasets and prediction horizons. On average, our model achieves a 15\% gain in performance compared to GPT4TS \citep{zhou2023gpt4ts}, which is a recent LLM based on GPT2. Additionally, TSMamba outperforms the state-of-the-art task-specific time-series model PatchTST \citep{patchtst}.
	
	\section{Conclusions}
	
	This paper introduces a Mamba-based foundation model for time series forecasting. The model employs both forward and backward Mamba encoders to capture temporal dependencies. To meet the challenges of forecasting heterogeneous time series data across different domains and frequencies with limited training resources, it leverages knowledge from the Mamba language model and adapts to time series datasets through a two-stage transfer learning approach. While the foundation model processes each channel of multivariate time series independently, it can capture cross-channel dependencies during fine-tuning on specific datasets using an additional compressed cross-channel attention module. TSMamba performs comparably to state-of-the-art models in zero-shot and full-shot settings while requiring significantly less training data, underscoring its potential to enhance downstream forecasting tasks.

\bibliographystyle{iclr2025_conference}
\bibliography{myref}


\end{document}